\documentclass[
aps,%
12pt,%
final,%
notitlepage,%
oneside,%
onecolumn,%
nobibnotes,%
nofootinbib,
superscriptaddress,%
noshowpacs,%
centertags]%
{revtex4}
\pdfoutput=1

\begin{document}
	\selectlanguage{english} 
	\title{Optic Disc and Cup Segmentation Methods for Glaucoma Detection with Modification of U-Net Convolutional Neural Network}
	\author{\firstname{Artem} \surname{Sevastopolsky}}
	\email[]{artem.sevastopolsky@gmail.com}
	\affiliation{Department of Mathematical Methods of Forecasting, Faculty of Computational Mathematics and Cybernetics, \linebreak Lomonosov Moscow State University}
	%
	%
	
	\begin{abstract}
		Glaucoma is the second leading cause of blindness all over the world, with approximately 60 million cases reported worldwide in 2010. If undiagnosed in time, glaucoma causes irreversible damage to the optic nerve leading to blindness. The optic nerve head examination, which involves measurement of cup-to-disc ratio, is considered one of the most valuable methods of structural diagnosis of the disease. Estimation of cup-to-disc ratio requires segmentation of optic disc and optic cup on eye fundus images and can be performed by modern computer vision algorithms. This work presents universal approach for automatic optic disc and cup segmentation, which is based on deep learning, namely, modification of U-Net convolutional neural network. Our experiments include comparison with the best known methods on publicly available databases DRIONS-DB, RIM-ONE v.3, DRISHTI-GS. For both optic disc and cup segmentation, our method achieves quality comparable to current state-of-the-art methods, outperforming them in terms of the prediction time.
	\end{abstract}
	\maketitle
	\section{Introduction}
	
	Glaucoma is the second leading cause of blindness all over the world, with approximately 60 million cases reported worldwide in 2010, and an increase by 20 million is expected in 2020 \cite{med_almazroa, med_quigley}. If left unnoticed, glaucoma can cause irreversible damage to the optic nerve leading to blindness. Therefore, diagnosing glaucoma at early stages is very important \cite{med_almazroa}. 
	
\begin{figure*}[h!]
		\setcaptionmargin{2mm}
		\onelinecaptionsfalse 
		\vspace{0.5cm}
		\subfigure[Healthy eye]
		{
			\includegraphics[width=0.23\linewidth]{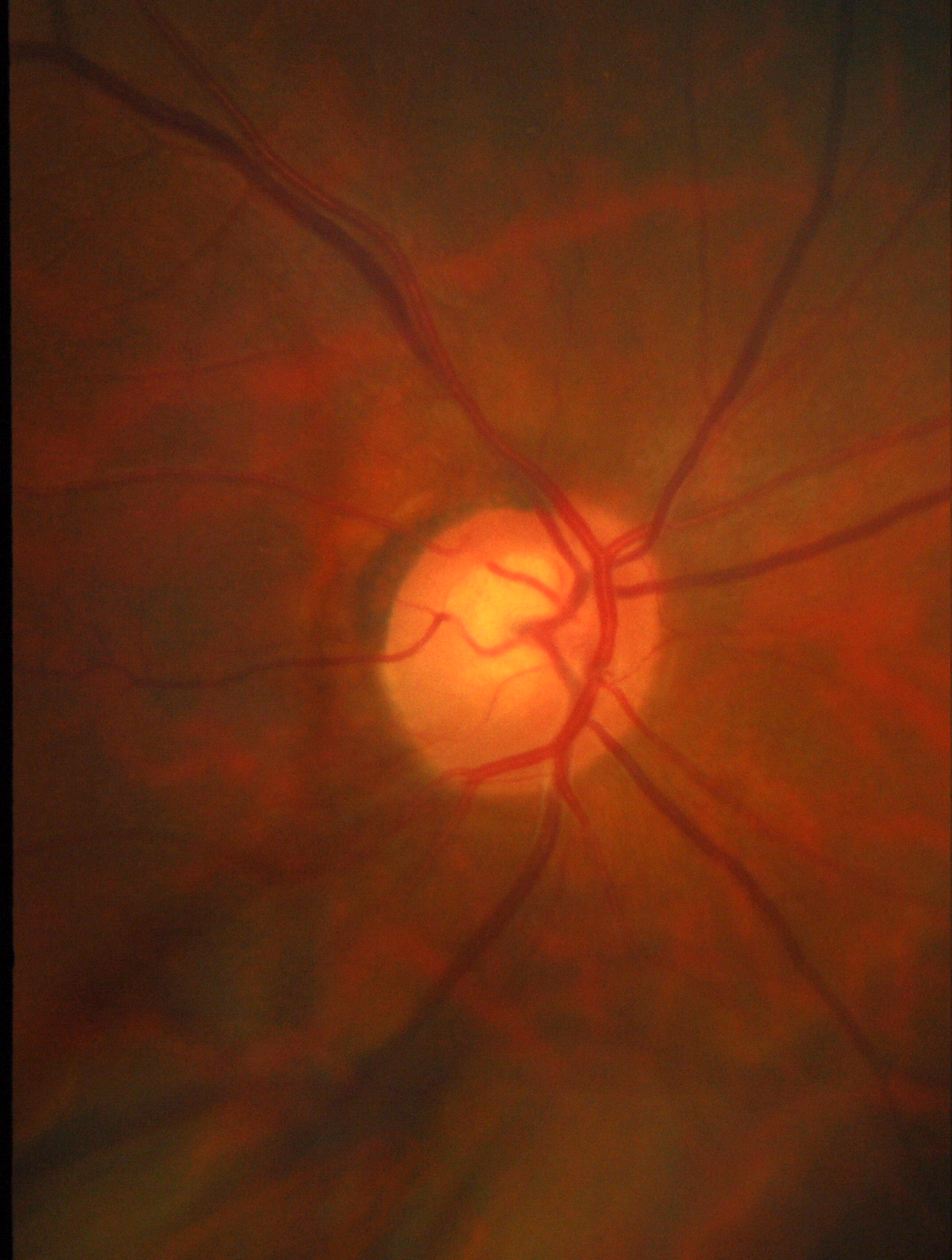}
			\includegraphics[width=0.23\linewidth]{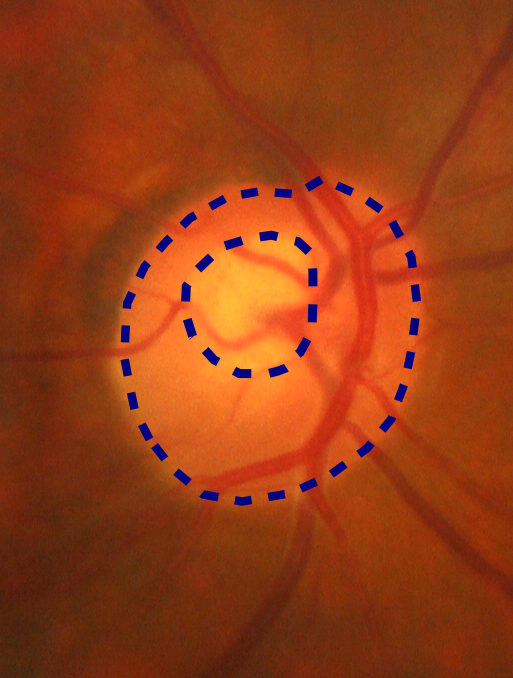}
		}
		\hspace{0.1cm}
		\subfigure[Glaucoma-suspicious eye]
		{
			\includegraphics[width=0.229\linewidth]{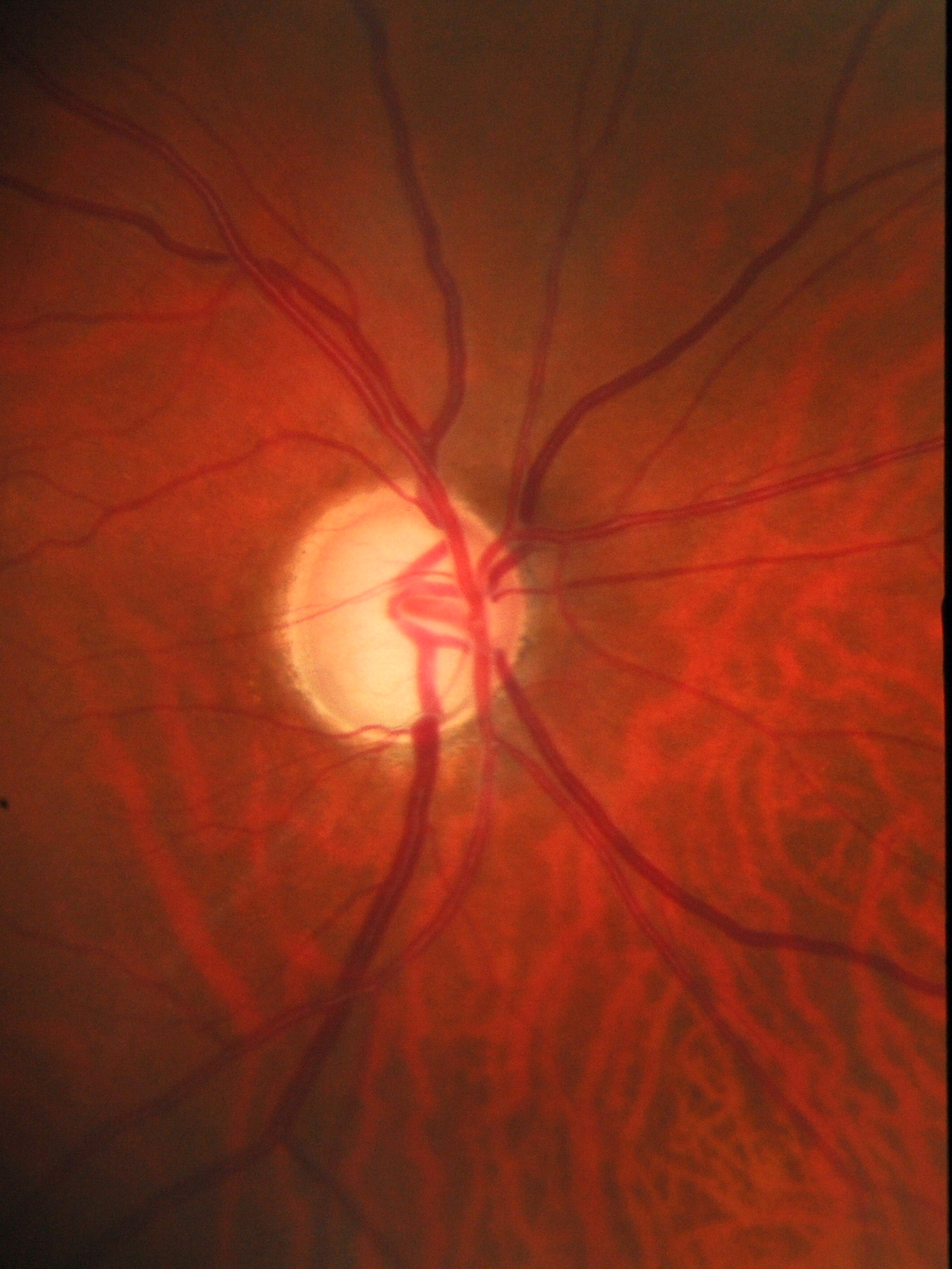} 
			\includegraphics[width=0.229\linewidth]{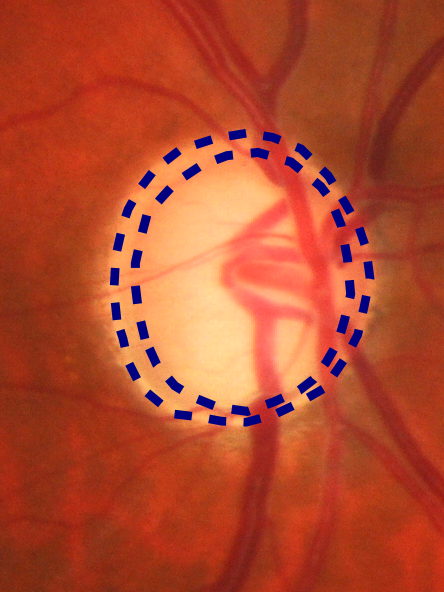} 
		}
		\captionstyle{normal}
		\caption{An example of healthy and glaucoma-suspicious eye from RIM-ONE v.3 \cite{rim_one_v3} database. Right-hand picture of each example contains enlarged optic disc area, where optic disc border is indicated by outer dashed line, optic cup border --- by inner dashed line. Note that CDR is larger for glaucoma-suspicious eye.}
		\label{fig:eyes_example}
	\end{figure*}

	Optic nerve examination includes eye fundus test, which requires a doctor localizing areas of optic disc and optic cup (central part of optic disc) and finding their borders. Presence of glaucoma can be identified by noticing optic nerve cupping, i.e. increase of optic cup in size. One of the main indicators of the disease is cup-to-disc ratio (CDR) --- a ratio between heights of cup and disc \cite{med_almazroa}. It is considered one of the most representative features of optic disc and cup areas for glaucoma detection, and, according to \cite{akram}, eye with CDR of at least 0.65 is usually considered as glaucomatous in clinical practice. Fig.~\ref{fig:eyes_example} shows an example of healthy and glaucoma-suspicious eye.
	
	Segmentation of the optic disc and cup and determination of the CDR are very time-consuming tasks currently performed only by professionals. As stated in \cite{med_lim}, according to a research, full segmentation of optic disc and cup requires about eight minutes per eye for a skilled grader. Solutions for automated analysis and assessment of glaucoma can be very valuable in various situations, such as mass screening and medical care in countries with significant lack of qualified specialists \cite{bastawrous, lodhia}. 
	
	There are several approaches to development of computer vision algorithms for glaucoma detection based on eye fundus images. First approach is to determine the presence of the disease directly from fundus images, which involves either manual or automatic extraction of image features, derived from color, position and pairwise relation of pixels. Another approach is to build algorithms for optic disc and cup segmentation, then, based on that, read out disc and cup dimension and from that judge on presence of the disease. In this work we investigate the latter pipeline, since it can provide more transparent and reliable solution for a medical doctor.
	
	Recognition quality and prediction time are the major requirements to the solution for automatic segmentation of eye parts. In order for a computer to be a decision-making system or at least an automatic eye scanner, it must make segmentation errors very seldom. Prediction time is also very important, especially when it is required to analyze large number of pictures in a small amount of time. Training time may be a concern in case retraining of an algorithm on larger database is needed frequently. However, exact requirements to the method depend on a specific setting of an automatic assessment system.
	
	\section{Related work}
	
	In this section we give an overview of several methods for optic disc and cup segmentation that have been evaluated by their authors on publicly available datasets with both images and groundtruth provided. 
	
	For optic disc segmentation task, authors of \cite{driu} use Fully-convolutional neural network \cite{fcn} based on VGG-16 net \cite{vgg} and transfer learning technique. They achieve superhuman quality of recognition in terms of Dice score (see section~\ref{section:our_approach} of this paper) and boundary error (mean distance between the boundary of the result and that of the ground truth), since obtained results are more consistent with a gold standard than a second human annotator used as control.
	
	For optic cup segmentation task, authors of \cite{bcf} use 2-layer multi-scale convolutional neural network trained with boosting. Training process pipeline is multi-stage and includes patches preparation and neural network training. For pre-processing, entropy filtering \cite{gonzalez_woods} in L*a*b* color space is performed for extracting the most important points of an image, followed by contrast normalization and stardardization of patches. Gentle AdaBoost \cite{dogan_adaboost} algorithm is then used to train convolutional filters, which are represented as linear regressors for small patches. At the test time, image propagation through the network is followed by unsupervised graph cut \cite{salah_graphcut}. The method was evaluated on DRISHTI-GS \cite{drishti_gs_1, drishti_gs_2} database, and it outperformed all other existing methods in terms of Intersection-over-Union score and Dice score (see section~\ref{section:our_approach} of this paper). However, it is necessary to note that this method crops images by area of their optic disc (cup) before performing segmentation of the optic disc (cup). It makes the method not applicable to new, unseen images of full eye fundus, since it requires a bounding box of optic disc and cup to be available in advance.
	
	The paper \cite{zilly} suggests an improvement to the aforementioned method in the training procedure for convolutional filters. Evaluation on DRISHTI-GS and RIM-ONE v.3 \cite{rim_one_v3} databases for optic disc and cup is provided. Compared to the previous method, it does not require the images to be cropped by the area of optic cup for its segmentation, which makes the solution applicable to previously unseen images.
	
	Method from the paper \cite{driu} has several drawbacks. It uses a deep neural network which takes a long time to train, model is large in terms of size of the file with network parameters and amount of required GPU memory. Authors of the paper were not pursuing a goal of the optic cup segmentation, which is a more challenging task than the optic disc segmentation. Besides, we were unable to reproduce the reported results. Methods from \cite{bcf} and \cite{zilly} are very complicated, hard to program and to reproduce the results. Being prepared for execution on CPU, they also have large prediction time. As written before, \cite{bcf} method required images to be cropped by the area of optic cup in advance, which is another drawback of a method. Some methods that are not mentioned in this section, such as \cite{med_lim, huiqi_li, jose}, have mostly been evaluated either on datasets that are not currently publicly available, or on very small datasets, or used metrics dependent on proportion between classes, thus making it harder to compare with them.
	
	\section{The presented approach}
	\label{section:our_approach}
	
	In this section, the universal method is proposed for segmentation of optic disc and cup. Our approach is primarily based on deep learning techniques, which have made a revolution in all tasks of computer vision in the last years and currently provide state-of-the-art solutions in image classification, segmentation and many other image recognition tasks. Another advantage of convolutional neural networks as main tools of deep learning is their universality, as the same network can usually recognize various patterns in different images and for different objects. 
	
\begin{figure*}[h] 
		\centering
		\setcaptionmargin{2mm} 
		\onelinecaptionstrue 
		\vspace{0.3cm}
		\includegraphics[width=0.8\linewidth]{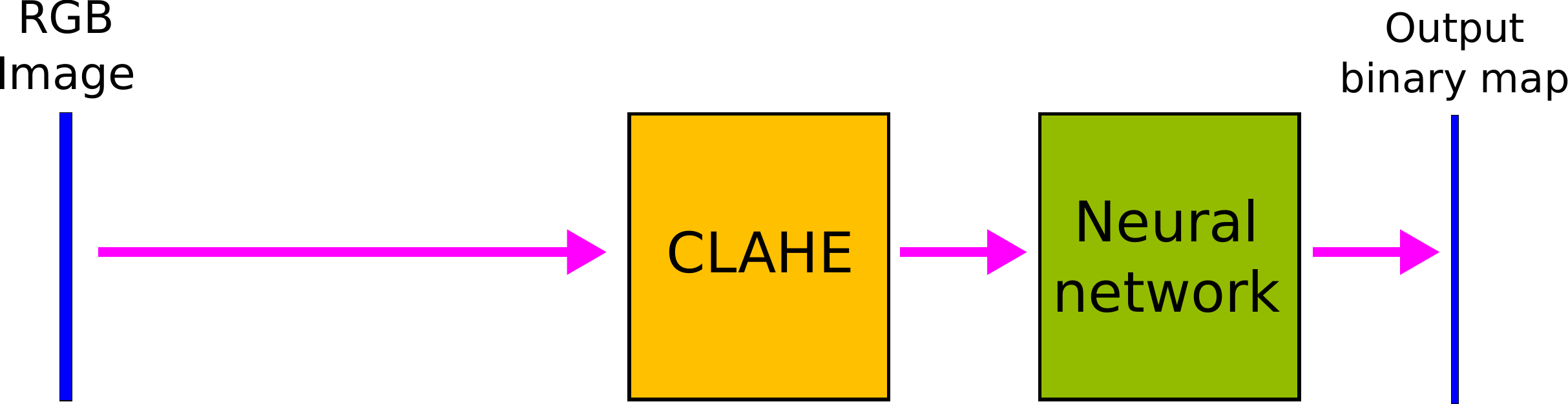} 
		\captionstyle{normal} 
		\caption{Pipeline of the proposed method for the task of optic disc segmentation.}
		\label{fig:optic_disc_pipeline}
	\end{figure*}

\begin{figure*}[h!]
		\setcaptionmargin{5mm} 
		\onelinecaptionstrue 
		\vspace{0.12cm}
		\includegraphics[width=0.8\linewidth]{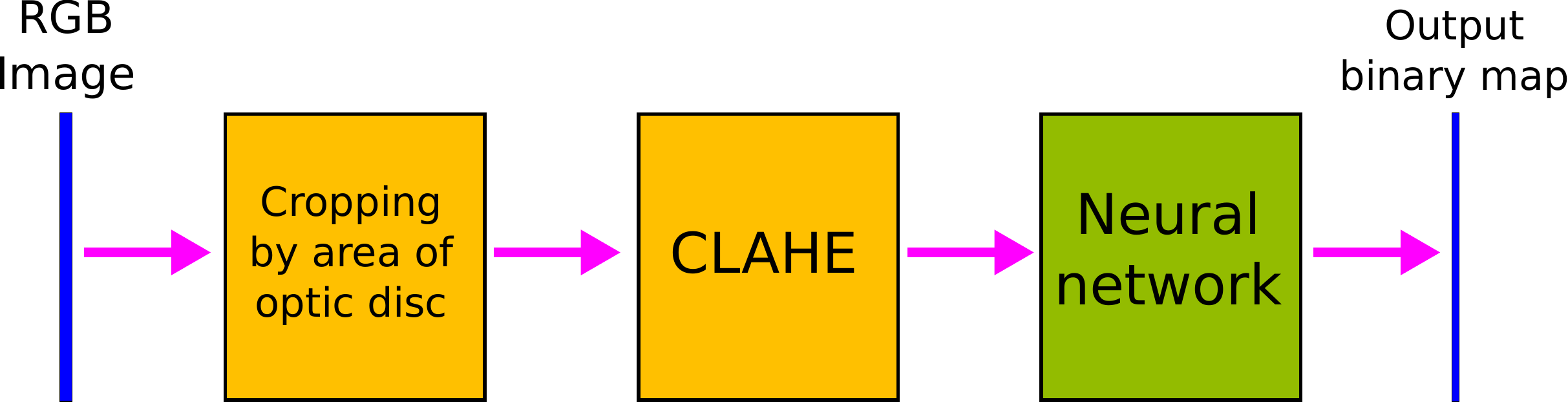} 
		\captionstyle{normal} 
		\caption{Pipeline of the proposed method for the task of optic cup segmentation.}
		\label{fig:optic_cup_pipeline}
	\end{figure*}

	Fig.~\ref{fig:optic_disc_pipeline} presents a pipeline of our method for optic disc segmentation, Fig.~\ref{fig:optic_cup_pipeline} --- for optic cup segmentation. Contrast Limited Adaptive Histogram Equalization (CLAHE) \cite{szeliski_clahe} is used as a pre-processing for both methods. It equalizes contrast by changing color of image regions and interpolating the result across them. For optic cup, we firstly crop the images by bounding box of optic disc (with margin from each side), which can be acquired from trained algorithm for optic disc.
	
	\normalsize

	Core component of the method is a convolutional neural network built upon U-Net \cite{u_net}. It is a neural network for image segmentation that accepts image as an input and returns probability map as an output. \mbox{U-Net} was introduced as a Fully-convolutional neural network capable of training on extremely small datasets and achieving results competitive with sliding-window based models. Trained with specific data augmentation and enhancement techniques, it outperforms existing methods on several biomedical image segmentation challenges \cite{u_net}. 
	
	\begin{figure*}[t!] 
		\setcaptionmargin{5mm} 
		\def\svgwidth{\linewidth}
		\fontsize{8}{5}\selectfont
		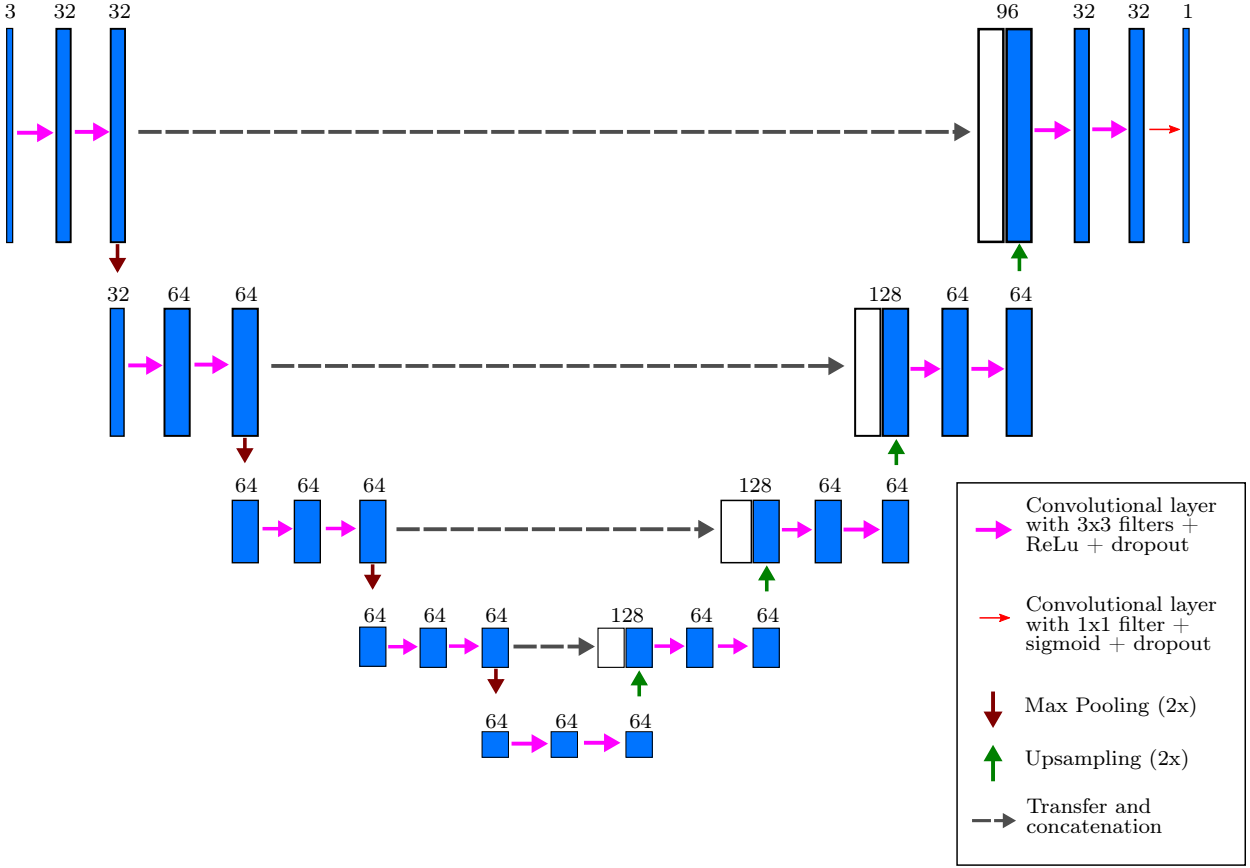
		\onelinecaptionstrue 
		\captionstyle{normal} 
		\caption{Architecture of neural network employed in our method.}
		\label{fig:our_u_net}
	\end{figure*} 
	\normalsize
	
	The architecture presented in the paper is depicted in Fig.~\ref{fig:our_u_net}. Like the original U-Net, it consists of contracting path (left side) and an expansive path (right side). Contracting path structurally repeats a typical architecture of convolutional part of the classification network, e.g. VGG-16 \cite{vgg}. On the expansive path, information is merged from layers of contracting path of appropriate resolution and layers of expansive path of lower resolution, so that a whole network recognizes patterns at several scales. Input image is firstly passed through a convolutional layer with filters of 3 x 3 pixels spatial resolution; number of filters in a layer is shown in the figure above a blue rectangle representing layer's output. Afterwards, Dropout regularization \cite{dropout} and ReLu activation function ($f(x) = max(0, x)$) are applied. The same is repeated again, and Max Pooling operation is applied, reducing image width and height by two. Image is then passed through aforementioned sequence of layers multiple times, until resolution is low enough. On the expansive path, the same convolutional layers are applied, interleaved with Upsampling layers, which raise image width and height by two in a trivial way.
	
	Compared to original \mbox{U-Net}, the presented modification has less filters in all convolutional layers and does not possess an increasing number of filters for decreasing resolution. Our experiments have shown that these changes do not lower quality of recognition for our tasks, but make the architecture much more lightweight in terms of number of parameters and training time. As a loss function, we use $l(A, B)$:
	\begin{gather*}
	l(A, B) = -\log d(A, B), \text{where:}\\
	d(A, B) = \frac{2\sum\limits_{i, j} a_{ij} b_{ij}}
	{\sum\limits_{i, j} a_{ij}^2 + \sum\limits_{i, j} b_{ij}^2},
	\end{gather*}
	
	where $A = (a_{ij})_{i=1\, j=1}^{H\;\;\;W}$ is a predicted output map, containing probabilities that each pixel belongs to the foreground, and $B = (b_{ij})_{i=1\, j=1}^{H\;\;\;W}$ is a correct binary output map. 
	
	$d(A, B)$ is an extension of Dice score for binary images $Dice(A, B) = \frac{2|A \cap B|}{|A| + |B|}$: if $A$ and $B$ contain only binary values, $d(A, B)$ and $Dice(A, B)$ are equal, but $d(A, B)$ also supports values that lie in $(0, 1)$. This extension allows us to compute gradient of the loss function. Stochastic Gradient Descent (SGD) with momentum \cite{sgd} was used as an optimization method. During the training, data augmentation was used to enlarge the training set by artificial examples. Images were subject to random rotations, zooms, shifts and flips.
	
	It is necessary to note that the proposed method does not require any preliminary cropping of input images to area of the optic disc, as it can segment the optic disc and the optic cup on a full eye fundus image. Detailed comparison of the presented method with the existing ones is given in the section~\ref{section:experiments}.
	
	\section{Experiments}
	\label{section:experiments}
	
	This section of the paper contains comparison between our solution and existing methods for both considered tasks. Results are reported for publicly available datasets DRIONS-DB \cite{drions_db}, RIM-ONE v.3 \cite{rim_one_v3}, DRISHTI-GS \cite{drishti_gs_1, drishti_gs_2}, which contain groundtruth segmentation for optic disc (and some for optic cup as well). DRIONS-DB contains 110 full eye fundus images with optic disc segmentation; RIM-ONE v.3 --- 159 images cropped by optic disc area, such that its diameter occupies about a fifth part of an image side length, with optic disc and cup segmentation; DRISHTI-GS --- 50 full eye fundus images with optic disc and cup segmentation. We evaluate the quality of trained algorithms by Intersection-over-Union (IOU) score: $\frac{|A \cap B|}{|A \cup B|}$ and Dice score: $\frac{2|A \cap B|}{|A| + |B|}$, where $A = (a_{ij})_{i=1\, j=1}^{H\;\;\;W}$ is a predicted output map, containing probabilities that each pixel belongs to the foreground, and $B = (b_{ij})_{i=1\, j=1}^{H\;\;\;W}$ is a correct binary output map. These quality measures do not depend on image scale, object scale and class imbalance. Dice score is also equal to $F_1$ score --- harmonic mean of precision and recall.
	
	We used a learning rate of $10^{-3}$ for optic disc and a learning rate of $3 \cdot 10^{-4}$ for optic cup segmentation. Momentum was set to 0.95, mini-batch of size 1 was used in order to minimize required amount of GPU memory. Resolution of input images was set to 256 x 256 for optic disc and to 512 x 512 for optic cup segmentation before their cropping. Region of interest was then resized to 128 x 128 by bilinear interpolation.
	
	For the task of optic disc segmentation, we compare our solution with the method from \cite{driu} paper (further referred as DRIU, as the name of the paper suggests), which is the best method that we have found in terms of IOU and Dice score functions for investigated datasets. For the task of optic cup segmentation, we compare with the method from \cite{bcf} (further referred as BCF, as the name of the paper suggests) and from \cite{zilly}. Score estimates are computed by cross-validation with 5 folds.
	
	\begin{table}[ht]
		\setcaptionmargin{5mm} 
		\onelinecaptionstrue 
		\captionstyle{flushleft} 
		\caption{Comparison of methods for optic disc segmentation. \glqq---\grqq\textrm{ }indicates that the result is not reported. Training time is computed as a product of one epoch time and average number of epochs.}
		\label{table:optic_disc_results}
		\bigskip
		\begin{tabular}{|l|c|c|c|c|c|c|c|}
			\hline
			& \multicolumn{2}{c|}{\,DRIONS-DB\,} &
			\multicolumn{2}{c|}{\,RIM-ONE v.3\,} &
			Training time on &
			Prediction &
			\# parameters\\
			& IOU & Dice & IOU & Dice & RIM-ONE v.3 & time & \\ \hline
			Our approach & \textbf{0.89} & 0.94 & \textbf{0.89 }& 0.95 & 26 s $\cdot$ 382 = 9932 s & \textbf{0.1 s} & $6,6 \cdot 10^5$ \\
			DRIU \cite{driu} & 0.88 & \textbf{0.97} & \textbf{0.89} & \textbf{0.96} & 56 s $\cdot$ 200 = 11200 s & 0.13 s & $1,5 \cdot 10^7$ \\
			Zilly et al. \cite{zilly} & --- & --- & \textbf{0.89} & 0.94 & \textbf{3296 s} &  5.3 s & \textbf{1890} \\ \hline
		\end{tabular}
	\end{table}
	\begin{table}[ht]
		\setcaptionmargin{5mm} 
		\onelinecaptionstrue 
		\captionstyle{flushleft} 
		\caption{Comparison of methods for optic cup segmentation. \glqq---\grqq\textrm{ }indicates that the result is not reported.}
		\label{table:optic_cup_results}
		\bigskip
		\begin{tabular}{|l|c|c|c|c|c|}
			\hline
			& \multicolumn{2}{c|}{\,DRISHTI-GS\,} &
			\multicolumn{2}{c|}{\,RIM-ONE v.3\,} &
			Prediction time \\   
			& IOU & Dice & IOU & Dice &  \\ \hline
			Our approach & 0.75 & 0.85 & 0.69 & \textbf{0.82} & \textbf{0.06 s} \\
			Zilly et al. \cite{zilly} & 0.85 & \textbf{0.87} & 0.80 & \textbf{0.82} & 5.3 s \\
			BCF \cite{bcf} & \textbf{0.86} & 0.83 & --- & --- & --- \\ \hline
		\end{tabular}
	\end{table}
	
	The presented algorithms were implemented on GPU with Python 2.7 programming language and Keras framework for training of neural networks (with Theano backend \cite{theano}). CLAHE implementation from Scikit-Image library was also used. All estimates of computational time are given for Amazon Web Services \cite{aws} g2.2xlarge instance with one NVIDIA GRID (Kepler GK104) GPU and Intel Xeon E5-2670 CPU for 256 x 256 images; estimate of Zilly et al. \cite{zilly} method's prediction time is given for a 2.66 GHz quad-core CPU, as reported. Prediction time of BCF \cite{bcf} is expected to be close to Zilly et al. \cite{zilly} prediction time, since these methods are very similar.
	
	\begin{figure*}[h!]
		\setcaptionmargin{2mm}
		\onelinecaptionsfalse 
		\vspace{0.5cm}
		\subfigure[Input image]
		{
			\includegraphics[width=0.14\linewidth]{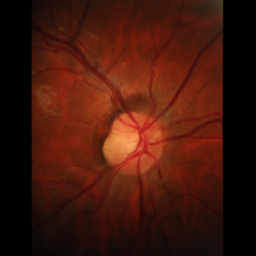} 
		}
		\subfigure[Predicted]
		{
			\includegraphics[width=0.14\linewidth]{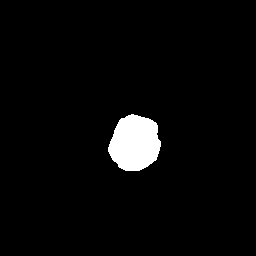} 
		}
		\subfigure[Correct]
		{
			\includegraphics[width=0.14\linewidth]{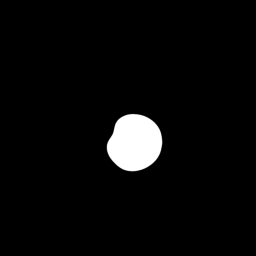} 
		}
		\hspace{0.4cm}
		\subfigure[Input image]
		{
			\includegraphics[width=0.14\linewidth]{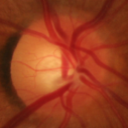} 
		}
		\subfigure[Predicted]
		{
			\includegraphics[width=0.14\linewidth]{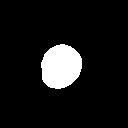} 
		}
		\subfigure[Correct]
		{
			\includegraphics[width=0.14\linewidth]{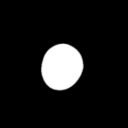} 
		}
		
		\subfigure[Input image]
		{
			\includegraphics[width=0.14\linewidth]{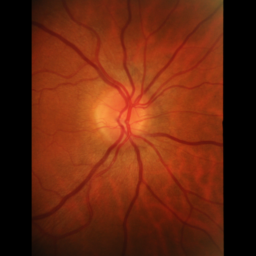} 
		}
		\subfigure[Predicted]
		{
			\includegraphics[width=0.14\linewidth]{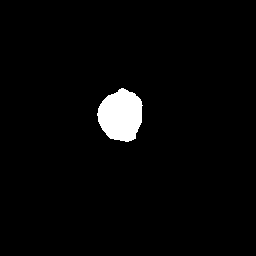} 
		}
		\subfigure[Correct]
		{
			\includegraphics[width=0.14\linewidth]{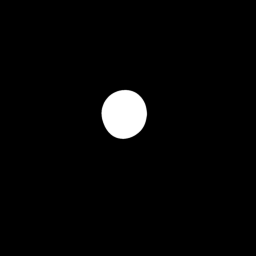} 
		}
		\hspace{0.4cm}
		\subfigure[Input image]
		{
			\includegraphics[width=0.14\linewidth]{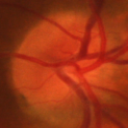} 
		}
		\subfigure[Predicted]
		{
			\includegraphics[width=0.14\linewidth]{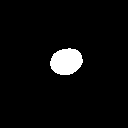} 
		}
		\subfigure[Correct]
		{
			\includegraphics[width=0.14\linewidth]{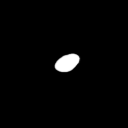} 
		}
		
		\captionstyle{normal}
		\caption{Visual comparison of the predicted results and correct segmentation on RIM-ONE v.3 for the optic disc (a)-(c), (g)-(i) and cup (d)-(f), (j)-(l). On (d)-(f), (j)-(l) region of the optic disc is shown as an input image. \newline For optic disc: (a)--(c): best case (IOU = 0.93, Dice = 0.97), (g)--(i): worst case (IOU = 0.80, Dice = 0.90); for optic cup: (d)--(f): best case (IOU = 0.93, Dice = 0.97), (j)-(l): worst case (IOU = 0.46, Dice = 0.64).}
		\label{fig:visual_results}
	\end{figure*}
	
	The results of the experiments indicate that the proposed method not only demonstrates quality competitive with quality of the existing methods in a majority of score metrics, but also possesses lowest prediction time, lowest training time among deep learning solutions, has small number of parameters (whole model can be saved in a file of only 5 MB; DRIU model requires about 120 MB) and is very easy to program with the use of modern frameworks. Despite that we gave estimates of prediction time for a machine equipped with modern (though not top level) GPU, for GPU with lower performance a prediction time can be only a few times larger. These advantages make the proposed method being a good solution for automatic glaucoma assessment on mobile devices.
	
	\section{Conclusion}
	In this paper we show that our method based on modified U-Net neural network can provide results similar or better than existing methods for the tasks of optic disc and cup segmentation on eye fundus images. The same method, applied to both tasks, achieves high quality of segmentation, which proves its applicability to various problems of image recognition. Advantages of the proposed solution also include its simplicity, simple programming with the use of modern frameworks and lowest possible prediction time. Experiments results and visual comparison show that automatic optic disc segmentation can be done at the quality competitive with human. However, optic cup is more challenging to recognize, which is supported by the fact that its border is much more subtle. We believe that there is a room for improvement for optic cup segmentation, and further research is needed.

	\begin{acknowledgments}
		We are especially grateful to Alexander~G.~D'yakonov, Professor, Dr. Sci. (Lomonosov MSU), for supporting and supervising this work. We would like to thank Leonid~M.~Mestetskii, Professor, Dr. Tech. (Lomonosov MSU), for initiating and supporting opthalmological research at the department. We are also grateful to Youth Laboratories company and especially to Konstantin Kiselev for provided computational resources.
	\end{acknowledgments}
	
	\newpage
	\section*{References}
	\bibliographystyle{ieeetr}
	\bibliography{paper}

\begin{thebibliography}{10}

\bibitem{med_almazroa}
A.~Almazroa, R.~Burman, K.~Raahemifar, and V.~Lakshminarayanan, ``Optic disc
  and optic cup segmentation methodologies for glaucoma image detection: a
  survey,'' {\em Journal of ophthalmology}, vol.~2015, 2015.

\bibitem{med_quigley}
H.~A. Quigley and A.~T. Broman, ``The number of people with glaucoma worldwide
  in 2010 and 2020,'' {\em British journal of ophthalmology}, vol.~90, no.~3,
  pp.~262--267, 2006.

\bibitem{akram}
M.~U. Akram, A.~Tariq, S.~Khalid, M.~Y. Javed, S.~Abbas, and U.~U. Yasin,
  ``Glaucoma detection using novel optic disc localization, hybrid feature set
  and classification techniques,'' {\em Australasian physical \& engineering
  sciences in medicine}, vol.~38, no.~4, pp.~643--655, 2015.

\bibitem{rim_one_v3}
F.~Fumero, S.~Alay{\'o}n, J.~Sanchez, J.~Sigut, and M.~Gonzalez-Hernandez,
  ``Rim-one: An open retinal image database for optic nerve evaluation,'' in
  {\em Computer-Based Medical Systems (CBMS), 2011 24th International Symposium
  on}, pp.~1--6, IEEE, 2011.

\bibitem{med_lim}
G.~Lim, Y.~Cheng, W.~Hsu, and M.~L. Lee, ``Integrated optic disc and cup
  segmentation with deep learning,'' in {\em Tools with Artificial Intelligence
  (ICTAI), 2015 IEEE 27th International Conference on}, pp.~162--169, IEEE,
  2015.

\bibitem{bastawrous}
A.~Bastawrous, H.~K. Rono, I.~A. Livingstone, H.~A. Weiss, S.~Jordan, H.~Kuper,
  and M.~J. Burton, ``Development and validation of a smartphone-based visual
  acuity test (peek acuity) for clinical practice and community-based
  fieldwork,'' {\em JAMA ophthalmology}, vol.~133, no.~8, pp.~930--937, 2015.

\bibitem{lodhia}
V.~Lodhia, S.~Karanja, S.~Lees, and A.~Bastawrous, ``Acceptability, usability,
  and views on deployment of peek, a mobile phone mhealth intervention for eye
  care in kenya: Qualitative study,'' {\em JMIR mHealth and uHealth}, vol.~4,
  no.~2, 2016.

\bibitem{driu}
K.-K. Maninis, J.~Pont-Tuset, P.~Arbel{\'a}ez, and L.~Van~Gool, ``Deep retinal
  image understanding,'' in {\em International Conference on Medical Image
  Computing and Computer-Assisted Intervention}, pp.~140--148, Springer, 2016.

\bibitem{fcn}
J.~Long, E.~Shelhamer, and T.~Darrell, ``Fully convolutional networks for
  semantic segmentation,'' in {\em Proceedings of the IEEE Conference on
  Computer Vision and Pattern Recognition}, pp.~3431--3440, 2015.

\bibitem{vgg}
K.~Simonyan and A.~Zisserman, ``Very deep convolutional networks for
  large-scale image recognition,'' {\em arXiv preprint arXiv:1409.1556}, 2014.

\bibitem{bcf}
J.~G. Zilly, J.~M. Buhmann, and D.~Mahapatra, ``Boosting convolutional filters
  with entropy sampling for optic cup and disc image segmentation from fundus
  images,'' in {\em International Workshop on Machine Learning in Medical
  Imaging}, pp.~136--143, Springer, 2015.

\bibitem{gonzalez_woods}
R.~Gonzalez, R.~Woods, and S.~Eddins, {\em Digital Image Processing Using
  MATLAB}.
\newblock Prentice-Hall, Inc., Upper Saddle River, NJ, USA, 2003.

\bibitem{dogan_adaboost}
H.~Do{\u{g}}an and O.~Akay, ``Using adaboost classifiers in a hierarchical
  framework for classifying surface images of marble slabs,'' {\em Expert
  Systems with Applications}, vol.~37, no.~12, pp.~8814--8821, 2010.

\bibitem{salah_graphcut}
M.~B. Salah, A.~Mitiche, and I.~B. Ayed, ``Multiregion image segmentation by
  parametric kernel graph cuts,'' {\em IEEE Transactions on Image Processing},
  vol.~20, no.~2, pp.~545--557, 2011.

\bibitem{drishti_gs_1}
J.~Sivaswamy, S.~Krishnadas, A.~Chakravarty, G.~Joshi, A.~S. Tabish, {\em
  et~al.}, ``A comprehensive retinal image dataset for the assessment of
  glaucoma from the optic nerve head analysis,'' {\em JSM Biomedical Imaging
  Data Papers}, vol.~2, no.~1, 2015.

\bibitem{drishti_gs_2}
J.~Sivaswamy, S.~Krishnadas, G.~D. Joshi, M.~Jain, and A.~U.~S. Tabish,
  ``Drishti-gs: Retinal image dataset for optic nerve head (onh)
  segmentation,'' in {\em Biomedical Imaging (ISBI), 2014 IEEE 11th
  International Symposium on}, pp.~53--56, IEEE, 2014.

\bibitem{zilly}
J.~Zilly, J.~M. Buhmann, and D.~Mahapatra, ``Glaucoma detection using entropy
  sampling and ensemble learning for automatic optic cup and disc
  segmentation,'' {\em Computerized Medical Imaging and Graphics}, vol.~55,
  pp.~28--41, 2017.

\bibitem{huiqi_li}
H.~Li and O.~Chutatape, ``Automated feature extraction in color retinal images
  by a model based approach,'' {\em IEEE Transactions on biomedical
  engineering}, vol.~51, no.~2, pp.~246--254, 2004.

\bibitem{jose}
J.~Jose and J.~Kuruvilla, ``Detection of red lesions and hard exudates in color
  fundus images,'' {\em International Journal of Engineering and Computer
  Science}, vol.~3, no.~10, pp.~8583--8588, 2014.

\bibitem{szeliski_clahe}
R.~Szeliski, {\em Computer vision: algorithms and applications}.
\newblock Springer Science \& Business Media, 2010.

\bibitem{u_net}
O.~Ronneberger, P.~Fischer, and T.~Brox, ``U-net: Convolutional networks for
  biomedical image segmentation,'' in {\em International Conference on Medical
  Image Computing and Computer-Assisted Intervention}, pp.~234--241, Springer,
  2015.

\bibitem{dropout}
G.~E. Hinton, N.~Srivastava, A.~Krizhevsky, I.~Sutskever, and R.~R.
  Salakhutdinov, ``Improving neural networks by preventing co-adaptation of
  feature detectors,'' {\em arXiv preprint arXiv:1207.0580}, 2012.

\bibitem{sgd}
I.~Sutskever, J.~Martens, G.~E. Dahl, and G.~E. Hinton, ``On the importance of
  initialization and momentum in deep learning.,'' {\em ICML (3)}, vol.~28,
  pp.~1139--1147, 2013.

\bibitem{drions_db}
E.~J. Carmona, M.~Rinc{\'o}n, J.~Garc{\'\i}a-Feijo{\'o}, and J.~M.
  Mart{\'\i}nez-de-la Casa, ``Identification of the optic nerve head with
  genetic algorithms,'' {\em Artificial Intelligence in Medicine}, vol.~43,
  no.~3, pp.~243--259, 2008.

\bibitem{theano}
{Theano Development Team}, ``{Theano: A {Python} framework for fast computation
  of mathematical expressions},'' {\em arXiv e-prints}, vol.~abs/1605.02688,
  May 2016.

\bibitem{aws}
``Amazon web services.'' https://aws.amazon.com.

\end{thebibliography}
	
	\section*{Authors}
	\begin{wrapfigure}{l}{25mm} 
		\includegraphics[width=1in,height=1.25in,clip,keepaspectratio]{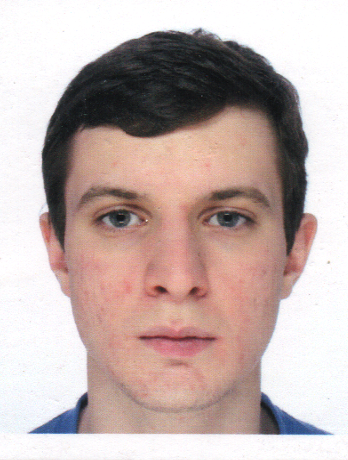}
	\end{wrapfigure}
	\textbf{Artem Sevastopolsky} (born in 1996) is a student of Lomonosov Moscow University, faculty of Computational Mathematics and Cybernetics, department of Mathematical Methods of Forecasting, graduating in 2017. His research interests include machine learning, computer vision, deep learning, image and video processing.\par
	
\end{document}